\documentclass[letterpaper]{article} 
\usepackage{aaai2026}  
\usepackage{times}  
\usepackage{helvet}  
\usepackage{courier}  
\usepackage[hyphens]{url}  
\usepackage{graphicx} 
\usepackage{natbib}  
\usepackage{caption} 
\usepackage{algorithm}
\usepackage{algorithmic}
\usepackage{amsmath} 
\usepackage{amsfonts} 
\usepackage{newfloat}
\usepackage{listings}
\DeclareCaptionStyle{ruled}{labelfont=normalfont,labelsep=colon,strut=off} 
\floatstyle{ruled}
\newfloat{listing}{tb}{lst}{}
\floatname{listing}{Listing}
\title{Multi-Horizon Time Series Forecasting of non-parametric CDFs with Deep Lattice Networks}
\author{
    Niklas Erdmann\textsuperscript{\rm 1},
    Lars Bentsen\textsuperscript{\rm 2},
    Roy Stenbro\textsuperscript{\rm 3}, 
    Heine Nygard Riise\textsuperscript{\rm 3}, 
    Narada Dilp Warakagoda\textsuperscript{\rm 1,4}, 
    Paal E. Engelstad\textsuperscript{\rm 1}
}
\affiliations{
    \textsuperscript{\rm 1}Department of Technology Systems, University of Oslo, 2007 Kjeller, Norway\\
    \textsuperscript{\rm 2}BearingPoint, 0252 Oslo, Norway\\
    \textsuperscript{\rm 3}Institute for Energy Technology, 2007 Kjeller, Norway\\
    \textsuperscript{\rm 4}Norwegian Defence Research Establishment, 2007 Kjeller, Norway
    
    \{niklase, n.d.warakagoda, paal.engelstad\}@its.uio.no, lars.bentsen@bearingpoint.com,  \{roy.stenbro, heine.riise\}ife.no
}

\newcommand{\citeauthorandyear}[1]{\citeauthor{#1} (\citeyear{#1})}

\begin{document}

\maketitle

\begin{abstract}
Probabilistic forecasting is not only a way to add more information to a prediction of the future, but it also builds on weaknesses in point prediction. Sudden changes in a time series can still be captured by a cumulative distribution function (CDF), while a point prediction is likely to miss it entirely. The modeling of CDFs within forecasts has historically been limited to parametric approaches, but due to recent advances, this no longer has to be the case. We aim to advance the fields of probabilistic forecasting and monotonic networks by connecting them and propose an approach that permits the forecasting of implicit, complete, and nonparametric CDFs. For this purpose, we propose an adaptation to deep lattice networks (DLN) for monotonically constrained simultaneous/implicit quantile regression in time series forecasting. Quantile regression usually produces quantile crossovers, which need to be prevented to achieve a legitimate CDF. By leveraging long short term memory units (LSTM) as the embedding layer, and spreading quantile inputs to all sub-lattices of a DLN with an extended output size, we can produce a multi-horizon forecast of an implicit CDF due to the monotonic constraintability of DLNs that prevent quantile crossovers. We compare and evaluate our approach's performance to relevant state of the art within the context of a highly relevant application of time series forecasting: Day-ahead, hourly forecasts of solar irradiance observations. Our experiments show that the adaptation of a DLN performs just as well or even better than an unconstrained approach. Further comparison of the adapted DLN against a scalable monotonic neural network shows that our approach performs better. With this adaptation of DLNs, we intend to create more interest and crossover investigations in techniques of monotonic neural networks and probabilistic forecasting.

\end{abstract}
\begin{links}
    \link{Code and Dataset}{https://github.com/Coopez/CDF-Forecasts-with-DLNs}
\end{links} 

\section{Introduction}
 
Probabilistic forecasting is a quickly growing field that aims to provide more information to a forecast than just a point estimate of the future (e.g. \cite{abdarReviewUncertaintyQuantification2021}). Forecasting finds application in many fields, including meteorology, economics, and energy production \cite{gneitingProbabilisticForecastsCalibration2007}.
In probabilistic forecasting, the future is predicted as a probability distribution. This can be used to give us more details about the future, such as the uncertainty or risk associated with a prediction. Forecasts about the future will always have a degree of uncertainty, which should be modeled \cite{priceProbabilisticWeatherForecasting2025}. In the energy market, for example, renewable energy prices are influenced by inherently variable weather conditions. Probabilistic forecasts can capture such variabilities and provide insights that can be used for informed risk management and bidding decisions by human or machine operators \cite{moralesElectricityMarketClearing2014}. These are processes in which optimization leads directly to a higher return in value and efficient use of resources (such as energy). 

Generally, probabilistic approaches can be categorized into parametric and non-parametric approaches. Parametric approaches assume a specific distribution of the data, in which prior knowledge about the data can be leveraged to reach better performance. Non-parametric approaches do not assume a specific distribution, but are also not bound to adhere to assumptions about an underlying distribution that may not hold true. 

Quantile regression (QR) \cite{koenkerRegressionQuantiles1978} is a non-parametric approach that is still actively developed and applied today. Due to its non-parametric nature, QR has a natural advantage in flexibility and adaptability compared to parametric approaches. It is also very much extendable for usage in time series forecasting \cite{wenMultiHorizonQuantileRecurrent2017}.  As a result of this, QR is widely used in areas like weather and energy forecasting \cite{songNoncrossingQuantileRegression2024,lauretProbabilisticSolarForecasting2017}.   
Recently, several novel advances have shown similar ways to use neural networks in combination with QR to implicitly model all quantiles between 0 and 1 continuously in a neural network's latent space \cite{tagasovskaSinglemodelUncertaintiesDeep2019a, dabneyImplicitQuantileNetworks2018,gasthausProbabilisticForecastingSpline2019}. 
Related work coined this technique as simultaneous quantile regression (SQR), or Implicit quantile networks (IQN). Here we will use SQR for brevity.
SQR enables us to implicitly estimate a non-parametric cumulative distribution function (CDF). By extension, it transforms a quantile regressor function into a complete inverse CDF of all quantile values. However, QR, and especially SQR (as shown previously in \citeauthorandyear{narayanExpectedPinballLoss2023}), is vulnerable to so-called quantile crossover, in which output values of lower quantiles cross over higher quantile values. Such a crossover decreases interpretability and applicability to real-world solutions. A modern car navigation system may provide three time-of-arrival estimates, such as: An optimistic 5 min, a likely 10 min, and a worst-case of 8 min. This does not does not make sense, as the worst-case time of arrival should not be shorter than the likely case. For our purposes, it also invalidates SQR, providing an implicit CDF, as it violates the monotonicity constraint of quantiles \cite{tagasovskaSinglemodelUncertaintiesDeep2019a}.

Quantile monotonicity must be guaranteed to ensure an accurate and legitimate approximation of a CDF. A proposed solution for quantile crossover is to develop a function approximator that can be monotonically constrained to the SQR quantile input. 
\citeauthorandyear{gasthausProbabilisticForecastingSpline2019} proposed to use splines in a quantile function with a recurrent neural network output (SQF-RNN). Their model can transform an input into parameters for a monotonic piecewise linear function of quantiles. Independently, \citeauthorandyear{brandoDeepNoncrossingQuantiles2022} implemented an approach whereby a positive neural network is considered the partial derivative of the conditional quantile function, resulting in a non-crossing inverse CDF.
\citeauthorandyear{narayanExpectedPinballLoss2023} proposed to use a Deep Lattice Network (DLN) for this task. Lattices are monotonically constrainable look-up tables \cite{guptaMonotonicCalibratedInterpolated2016b,youDeepLatticeNetworks2017}. \citeauthor{narayanExpectedPinballLoss2023} also compared their work to SQF-RNNs and found lattices to show higher accuracy and faster run times. SQF-RNNs have been applied to time series forecasting tasks before \cite{wangNonparametricProbabilisticForecasting2022}. However, DLN parametric complexity scales with the number of input features, which led to DLNs only being applied on problems with a limited number of input features \cite{narayanExpectedPinballLoss2023}. In particular, DLNs have never been used in time series forecasting.

There is ongoing development in preventing quantile crossover in QR with predefined quantile outputs. In the introduction of SQR, \citeauthorandyear{tagasovskaSinglemodelUncertaintiesDeep2019a} proposed a penalty function to prevent quantile crossover, which was built upon in future work \cite{guptaHowIncorporateMonotonicity2019,shenNonparametricEstimationNonCrossing2024a}. However, penalty functions do not provide a guarantee against crossovers.
 Providing such guarantees, \citeauthorandyear{songNoncrossingQuantileRegression2024} added a positive last layer to a QR-RNN to prevent quantile crossover in the context of a QR post-processing step for ensemble weather forecasts. Similarly, positive parameters are also used to prevent quantile crossover of QR in forecasting of wind power generation \cite{cuiEnsembleDeepLearningBased2023}. Unfortunately, positive parameter-based approaches are not transferable to SQR as they rely on a predefined set of quantiles.  
  Similarly, \citeauthorandyear{parkLearningQuantileFunctions2022} model a set of quantiles as an integral with non-negative elements to ensure monotonicity, and can use interpolation to extrapolate on non-trained quantiles. However, this does not model a complete CDF, but just approximates it.
  
It is also possible to simply sort the quantiles as a post-processing step \cite{chernozhukovQuantileProbabilityCurves2010a, smylAnyquantileProbabilisticForecasting2026}. While this can reportedly increase performance, it discards the trained quantile relations as part of the output of the function approximator.  

There also exist more general monotonic network approaches, competing and featuring different trade-offs compared to DLNs. Monotonic networks can guarantee monotonicity with respect to one (or more) input feature(s), which is all that is needed to prevent quantile crossovers in SQR. Certified monotonic neural networks (CMNNs) \cite{liuCertifiedMonotonicNeural2020} reported to offer better performance than DLNs, while constrained monotonic neural networks \cite{runjeConstrainedMonotonicNeural2023} were shown to beat CMNNs. Scalable monotonic neural networks (SMNNs) \cite{kimScalableMonotonicNeural2023} outperform CMNNs and DLNs. In their testing, SMNNs perform similarly to, or slightly better than, constrained monotonic neural networks. However, monotonic network comparisons do not take place in a time series forecasting context, which makes it difficult to know if they could and would work with a forecasting task. 

Numerous studies in probabilistic forecasting and monotonic networks have explored different ways that can be used to prevent quantile crossover. There is also significant interest in non-parametric probabilistic forecasting from an applied forecasting perspective. Comparatively, however, there is very little work on combining these probabilistic forecasting and monotonic networks in approaches that try to forecast an inverse CDF with a network that is guaranteed to be monotonic with respect to its quantile input. 

\subsection{Contributions}
In this paper, we propose an adaptation of a DLN to time series forecasting with SQR. This represents a missing link between probabilistic forecasting or non-parametric CDFs and monotonic constraintable networks. 

We test this novel approach on a solar irradiance forecasting task. Solar irradiance is a particularly relevant application of weather forecasting techniques to enable the day-ahead prediction of photovoltaic energy production (of which solar irradiance data is an excellent surrogate, e.g. \cite{antonanzasReviewPhotovoltaicPower2016}). The increasing prevalence of renewable energy sources and their natural, weather-related variations necessitate making predictions about them \cite{yangReviewSolarForecasting2022}. Accurate probabilistic forecasting is shown to play an increasingly important role in the general methodology. Ramps, high changes in observed irradiance due to varying cloud coverage, may only in some cases be possible to capture by point forecasters, but can be captured by a CDF of a probabilistic forecaster \cite{samuApplicationsSolarIrradiance2021}. Predictions taking uncertainty and risk into account enable more informed bidding decisions in the energy market and make photovoltaic production more economically efficient. By extension, probabilistic forecasting may facilitate even higher investment in this area \cite{moralesElectricityMarketClearing2014}. 

With this application, we show that we can not only connect existing fields in a novel way but also show that the resulting approaches may be directly applied in relevant applications. 
We hope that our work can facilitate further interaction between two largely independent research streams, applied forecasting and monotonic networks, which may lead to further improvements in the future. 
Specifically, the list of our contributions is as follows:
\begin{itemize}
    \item We adapt a DLN to time series forecasting with SQR, which is the first adaptation of a DLN to this task.
    \item We detail the implementation and special considerations needed to fit DLN to time series forecasting with SQR.
    \item We compare our flavor of DLN to popular solar irradiance forecasting approaches and a forecasting adaptation of SMNNs, which have been shown to outperform DLNs in other tasks. In these comparisons, our iteration of a DLN shows better results, highlighting the applicability of Lattices for use in probabilistic forecasting.  
\end{itemize}

\section{Methods}

\subsection{Forecasting Task}

In solar irradiance forecasting, the goal is to predict a horizon of future values given a window of past observations. This could be a useful forecasting application for owners of a photovoltaic power plant, looking to optimize their sales on the day-ahead energy market \cite{samuApplicationsSolarIrradiance2021}.
Formally, we define the forecasting task as given a series of $\chi$ past values containing $F$ features, predict a series of $\gamma$ future values. The past values are defined as a window $w$ of $\chi = (x_{t-w}, \dots, x_{t})\text{, with } \chi \in \mathbb{R}^{w \times F}$, while the future values are defined as a horizon $h$ of $\gamma = (y_{t+1}, \dots, y_{t+h})$ with $\gamma \in\mathbb{R}^{h\times 1}$. The trainable forecaster is a function $f$ mapping the window of past values to the horizon of future values, i.e. \begin{equation}\label{eq:forecaster}
\gamma = f(\chi)
\end{equation} 
 As a potential application to the day-ahead energy market, we choose a forecasting horizon $h$ of 36 hours and 96 as the past window $w$, i.e., 4 days. 
\subsection{Quantile Regression}
QR in time series forecasting (even if monotonically constrained) is usually trained to output a list of predictions containing a predetermined set of quantiles of interest, e.g., $[0.1,0.25,0.5,0.75,0.9]$.
This can be advantageous against parametric approaches, as we only learn a categorical subset of the underlying distribution. However, it also contains less information than parametric approaches, as the resulting categorical distribution is not a complete CDF.

In QR, we minimize the pinball loss $l_\tau(\gamma,\gamma')$, with $\gamma$ representing ground truth, and $\gamma '$ representing the prediction of a forecaster. Pinball loss is dependent on a quantile level $0 \leq \tau \leq 1$. The loss function looks as follows:
\begin{equation}\label{eq:pinball_loss}
l_\tau(\gamma,\gamma') = 
\begin{cases}
\tau(\gamma-\gamma') & \text{if\,} \gamma-\gamma' \geq 0, \\
(1-\tau)(\gamma'-\gamma) &  \text{else.}

\end{cases}
\end{equation}
Usually, we would now define a set of $\tau$-values and evaluate model output by iterating through and summarizing the pinball loss depending on $\tau$. 

SQR is an adaptation of default non-parametric quantile regression \cite{tagasovskaSinglemodelUncertaintiesDeep2019a, dabneyImplicitQuantileNetworks2018}.
Instead of predetermining a set of quantiles to learn, we add a random quantile to the input of a trainable function, such as a neural network, and learn to predict that. This approach enables a neural network to learn an implicit continuous quantile distribution.

With SQR, we draw $\tau \sim U(0,1)$ during training and use it as input into the model and loss function $f(\chi,\tau) = \gamma'$. Now we can update equation \ref{eq:forecaster} as predicting a probability distribution when aggregating a set of $T = (\tau_0,...\tau_q)$ with $q \in\mathbb{N}$ into a set of output values $Y \in \mathbb{R}^\tau$ shaping the inverse CDF :
\begin{equation}\label{eq:CDF_set}
 Y = (f(\chi,\tau_0),\dots,f(\chi,\tau_q))
\end{equation}
In training, we do not need to consider a set of $Y$, as we are drawing a different $\tau$ for every minibatch. Applying minibatches and combining equation \ref{eq:CDF_set} and equation \ref{eq:pinball_loss} we can thus optimize $f$ as follows:
\begin{equation}
arg \min_f \frac{1}{N} \sum_{i=1}^{N} E_{\tau_i \sim U(0,1)} l_\tau(f(\chi_i,\tau_i),\gamma_i)
\end{equation}
$N$ here being the batch size. 

Intuitively, $f$ is now an inverse and complete CDF with respect to $\tau$ \cite{narayanExpectedPinballLoss2023}. Nevertheless, QR via neural networks as described above does not protect against quantile crossovers, which are violations of the monotonicity constraint of quantiles. Quantiles of a probability distribution are monotonically increasing if, 
\begin{equation}\label{eq:monotonicity}
f(\chi,\tau_a) \leq f(\chi,\tau_b) \text{ for } \tau_a \leq \tau_b.
\end{equation}
A solution to quantile crossover is to use a function approximator that is monotonically constrainable with respect to (at least) one of its inputs, such as DLNs. Only by preventing quantile crossover can we produce a legitimate inverse CDF \cite{narayanExpectedPinballLoss2023}.
\subsection{DLNs}

DLNs are described as monotonically constrainable look-up tables \cite{guptaMonotonicCalibratedInterpolated2016b}.
A one-dimensional lattice function maps an input to an interpolated piecewise linear function. A higher-dimensional lattice function represents the same piecewise-linear function, but with a $ D$-dimensional input. $D$-lattices will have $k^D$ number of trainable parameters, where $k$ represents the knots, or keypoints, in the piece-wise linear function. In other words, a lattice is a hypercube, spanned by its input features, such that $[0,1]^D$.  
This property enables it to map all of its inputs to an output value, just as a lookup table would. It achieves a smooth and nonlinear function by interpolating inputs between its vertices. 
Lattices can be trained using backpropagation and are shown to be able to approximate any bounded, continuous function if they are supplied with enough knots in their piece-wise function.
In mathematical terms, a lattice maps $f(\chi):\mathbb{R}^D \to\mathbb{R}$ in which 
\begin{equation}\label{eq:lattice}
f_l(\chi) = \theta^T\psi(\chi)    
\end{equation}
 $\theta$ represents the values of the lattice and $\psi(x)$ the interpolation of the input over the interpolation weights \cite{narayanExpectedPinballLoss2023}. 
In \citeauthorandyear{guptaMonotonicCalibratedInterpolated2016b}, the interpolation step of a lattice is tested with a multilinear or a simplex approach. Here we only use multilinear. With $v_k \in [0,1]^D$ being the $k$th vertex of the unit hypercube, the interpolation weight on $v_k$ is shown as:
\begin{equation}
    \psi_k(\chi) = \prod^{D-1}_{d=0} \chi[d]^{v_k[d]}(1-\chi[d])^{1-v_k[d]}
\end{equation}
As $v_k$ is either 0 or 1, it determines that $\chi$ is multiplied either as $\chi[d]$ or $1-\chi[d]$. 
 
Intuitively, lattices can be applied similarly to regular neural networks, although featuring some prominent differences: A single lattice only ever outputs a scalar, even if the input is a vector (i.e., in practice, the output is a vector of the same size as the mini-batch size). As mentioned above, Lattices' complexity scaling is exponential with respect to the input dimensions $D$ ($k^D$). A large $D$ quickly becomes unfeasible to fit onto modern computers, which is why Lattices are usually applied to problems with a limited number of features.
One solution to this is to use an ensemble in which every lattice only takes a small subsample of the overall input features \cite{youDeepLatticeNetworks2017}. In this ensemble, low-dimensional lattices are initialized to a random subset of features and then trained.  

The originally proposed DLNs, which leverage such ensembles, can be built up of several distinct layers as described in detail by \citeauthorandyear{youDeepLatticeNetworks2017}.
\begin{itemize}
    \item \textbf{Constrainable linear embedding layer}: Featuring two matrices and one bias, which linearly embeds a monotonic input. One matrix is used for monotonic inputs, in which a monotonic constraint is applied by keeping its weights restricted to be non-negative. The other matrix is used for non-monotonic inputs. If splitting between monotonic weights $\theta^q\ge 0$ and non-monontonic weights $\theta^m$,
\begin{gather}\label{alg:constrained_lin}
\gamma = 
\begin{bmatrix}
\theta^q  \chi^q \\
\theta^m  \chi^m \\ 
\end{bmatrix}
+ b
\end{gather}
This has been used in conjunction with calibrators (described below) as a form of input layer.
\item \textbf{Calibration layer}: A layer of parallel one-dimensional lattice functions, embedding each feature separately. Any calibrator is a one-dimensional lattice with $k$ key-value pairs mapping input to output $(a \in\mathbb{R}^k, b \in\mathbb{R}^k)$. Two values of $b$ that correspond to $a$ values close to the input are interpolated to get the output. This layer is used as a preprocessing step or to give the DLN more expressiveness with respect to every feature.

\item\textbf{Lattice ensemble}: A set of lattices $S$ in which each lattice $f_l:\mathbb{R}^A\to\mathbb{R}, A \subset D$, in which all $A$ are unique with respect to each other $A \neq A'$. Outputs of this layer can be sent into another lattice ensemble until the output becomes scalar, be summarized, e.g., by averaging, or be sent through a final calibration step. 
\end{itemize}

Even though these layers are suggested to be used in a certain way in DLNs, they are all monotonically constrainable and may be stacked freely in any order.

\subsubsection{Monotonicity of DLNs}
In contrast to neural networks, lattice monotonicity constraints are relatively easy to enforce. As long as every pair of adjacent linearly interpolated look-up table parameters $\theta_s$ and $\theta_r$ obey $\theta_s > \theta_r$ in the same direction, then the lattice is monotonic \cite{guptaMonotonicCalibratedInterpolated2016b} (also see \citeauthorandyear{guptaMonotonicCalibratedInterpolated2016b} for the formal proof of monotonicity of lattices which applies the calibration layer and the lattice ensemble). All three different DLN layers, the constrainable linear, the calibration layer, and the lattice ensemble, preserve monotonicity across layers by treating the monotonic output from previous layers as monotonic input in subsequent layers \cite{youDeepLatticeNetworks2017}. In practice, lattice monotonicity is enforced by solving the constrained optimization with Dykstra's projection algorithm \cite{narayanExpectedPinballLoss2023}. 

\subsection{DLN adaptation to Time Series Forecasting}

Even with the changes to lattices introduced by DLNs, they cannot, out of the box, be applied to higher-dimensional forecasting problems. First, ensembles are not a perfect tool to shrink the complexity scaling of $k^D$. Lattices with a smaller $k$ or $D$ might be too small as they will be less expressive, or not be able to model many feature interactions. Too big lattices will run into a complexity bottleneck.
Second, DLNs' scalar output size limits architectural freedom when needing to meet a predefined horizon of forecasts. One lattice function only maps to one scalar output.

Fortunately, we do not need a DLN to interpret time series data; we only need it to produce an implicit CDF of the forecast. As such, we can embed higher-dimensional data into lower dimensions, with e.g. a neural network, and use that as input to a DLN. This idea was first discussed in \citeauthorandyear{narayanExpectedPinballLoss2023}, but never actually tested. Long-short-term memory (LSTM) models are very popular and perform relatively well in the solar irradiance forecasting (e.g. \cite{qingHourlyDayaheadSolar2018,srivastavaComparativeStudyLSTM2018,huangHybridDeepNeural2021})
For this work, we select an LSTM as a trainable embedding function for the time series data. This is an interchangeable decision, so more sophisticated models like Deep-AR \cite{salinasDeepARProbabilisticForecasting2020a}, or Transformers \cite{vaswaniAttentionAllYou2017} may result in even better forecasts. However, LSTMs show great performance in forecasting tasks and are a good experimental baseline for our intended comparison of showing that DLNs can be used to accurately forecast implicit CDFs. 
In addition to an LSTM, an ensemble of lattices is used to minimize the input-dependent size of the DLN. 

Conversely, fitting the output of a DLN to a predefined horizon can be achieved by employing a set of parallel lattices that take the same input and produce a different output timestep, or using the constrained linear embedding proposed by \citeauthorandyear{youDeepLatticeNetworks2017} as a constrained linear output layer. The latter is much preferred, as a linear output layer also permits us to reduce the usage of high-dimensional lattices to a minimum and only use a single layer of ensembles. 

\begin{figure}[ht]
\centering
\includegraphics[width=\columnwidth]{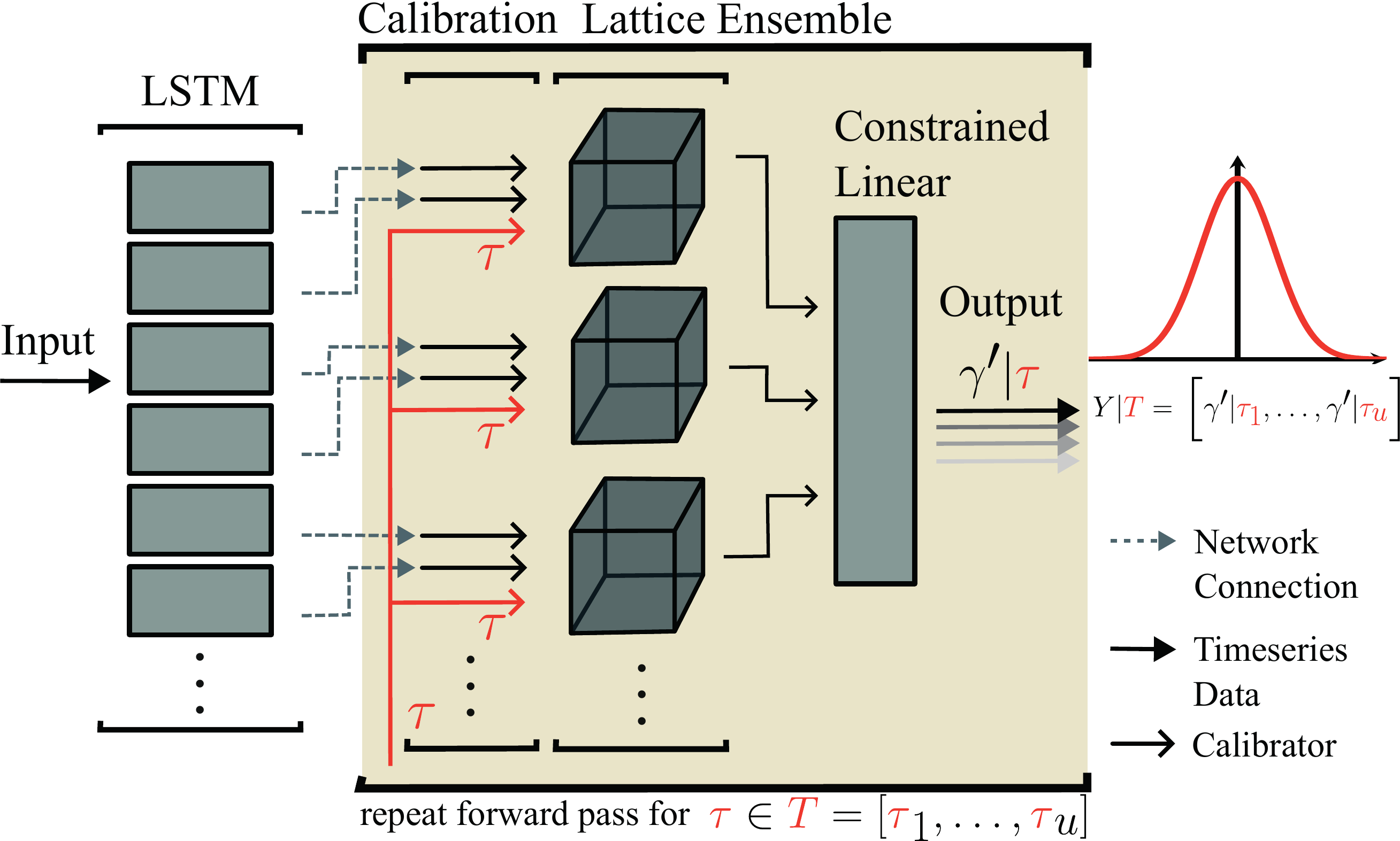}
\caption{Architecture of the LSTM-DLN for SQR. If the DLN is looped over a set of quantiles, a correspondingly detailed approximation of a CDF can be accumulated.}
\label{fig:dln}
\end{figure}

Finally, we change the quantile input as proposed by \citeauthorandyear{narayanExpectedPinballLoss2023}. Instead of feeding quantiles into only one lattice in the ensemble and letting constraints be propagated throughout the ensemble, we reserve one input feature in every lattice in the ensemble for quantiles. This way, every feature receives a directly trained relation to the implicit, complete CDF. Additionally, calibrators are added before and after the DLN layer.
The resulting architecture is shown in figure \ref{fig:dln}. 
We refer to the LSTM as the input model $f_1$ and the DLN as the output model $f_2$. $f_2$ needs one forward pass to model one quantile of the complete CDF. During training, $f_1$ and $f_2$ are called in sequence for each randomly drawn quantile, as can be seen in algorithm \ref{alg:training}. When desired (for example, during an exploitation step), $f_2$ can also be run for every quantile of interest, e.g., to form a highly detailed approximation of a CDF. This is shown in figure \ref{fig:dln} and algorithm \ref{alg:exploitation}.

\begin{algorithm}[tb]
\caption{Training of the proposed SQR Architecture.}
\label{alg:training}
\textbf{Input}: Collection of minibatches of past and target, $(X,\Gamma)$\\
\textbf{Output}: Trained LSTM $f_1$ and DLN $ f_2$\\
\begin{algorithmic}[1] 
\STATE Initialize $f_1$ and $ f_2$
\FOR{$e \leftarrow 1$ to $E$}
\STATE Sample $\chi_j,\gamma_j$ from $(X,\Gamma)$
\STATE Sample $\tau_j$ from $\mathcal{U}(0,1)$
\STATE Obtain embedded values $B$ from $f_1(\chi_j)$
\STATE Obtain ${\gamma'}^{\tau}_{j}$ from $f_2(B,\tau_j)$
\STATE Compute loss $L(\gamma_j,{\gamma'}^\tau_j,\tau_j)$ with eq. \ref{eq:pinball_loss}
\STATE Apply backpropagation to $f_1$ and $ f_2$ with $L$
\ENDFOR
\STATE \textbf{return} $f_1$ and $ f_2$
\end{algorithmic}
\end{algorithm}

\begin{algorithm}[tb]
\caption{Exploitation step of the proposed SQR Architecture.}
\label{alg:exploitation}
\textbf{Input}: Time series of past values $\chi$ and a set of quantiles 
$T=(\tau_0,\dots,\tau_q)$\\
\textbf{Output}: Set of forecasts $Y = (f(\chi, \tau_0),\dots, f(\chi, \tau_q) )$ \\
\begin{algorithmic}[1]
\STATE Initialize empty set $Y$
\STATE $B\leftarrow f_1(\chi)$
\FOR{ $\tau$ in $T$}
\STATE $Y_\tau \leftarrow f_2(B,\tau)$
\ENDFOR
\STATE \textbf{return} $Y$
\end{algorithmic}
\end{algorithm}

\subsection{Dataset}
The solar irradiance dataset consists of five years (2016-2020) of hourly surface downwelling shortwave radiation measurements from 20 stations around southern Norway\footnote{\url{https://thredds.met.no/thredds/projects/sunpoint.html}}. As additional features, satellite-based irradiance estimates (from CAMS-RAD\footnote{\url{https://www.soda-pro.com/web-services/radiation/cams-radiation-service}}) and estimated weather features (from NORA3\footnote{\url{https://thredds.met.no/thredds/projects/nora3.html}}) from the coordinates of all 20 sites are added. The weather data taken from NORA3 consists of integrated surface net downward short-wave flux, relative humidity, wind speed and direction, snowfall, precipitation, temperature, cloud area fraction, and surface air pressure. A clear sky index, indicating irradiance on a cloudless day, is added from CAMS McClear\footnote{\url{https://www.soda-pro.com/web-services/radiation/cams-mcclear}}. Lastly, time features for hour of day, day of week, and week of year are embedded in a circular sine and cosine component and added to the dataset. 
Overall, this results in 12 features per 20 stations with an additional 6 embedded time features, adding up to 246.
 One station (station 11, near the west coast of Norway) is chosen as the target for the forecaster, as it has the least amount of missing data. Other missing data were mostly night values and were replaced with zeroes.
 Norway experiences much longer days with much higher irradiance amplitudes in summer than in winter, which models will have to adapt to. Night values are often removed from the dataset as irradiance is zero. However, as the window of historical values used for a forecast is fixed and longer than a day, we need these values to keep the relative days of available data consistent. 

\subsubsection{Data Preprocessing}

The respective data sources provide cleaned data sets, so only a min-max scaling is applied to all features. Additionally, the dataset is split into a training, validation, and testing set. 
Testing and validation received one year each (2016 and 2017, respectively) to be able to infer performance on a whole year of data. 2017 and 2016 were the least correlated to the rest (2018-2020, which was used as a training set) to increase forecast difficulty.
 
\subsection{Evaluation}
We evaluate our adapted DLN on a set of 11 equidistant quantiles within the interval: $[0.025,0.975]$. Training and Hyperparameter optimization are done on the training and validation data, while final performance testing is done on the test set.

\subsubsection{Comparisons}
We designed the adapted DLN to be able to provide more information about a forecast by forecasting an implicit complete CDF. We consider this successful if we can manage to achieve similar performance to unconstrained versions of the same general architecture. This is the case, as unconstrained architectures can be expected to be more flexible than constrained ones, thus should return better results if the architecture can return well-calibrated CDFs. So, we are interested in testing only different versions of $f_2$ while $f_1$ is retrained, but with hyperparameters kept constant across different $f_2$.

We hypothesize that SQR requires a nonlinear model in the output. Thus, the first comparison is made with a linear and a constrained linear embedding layer as a singular output layer. 
Next, we need to test our constrained approach against an unconstrained neural network to see if our choice in architecture mitigates the decline in accuracy due to constraints.
Further, we identified SMNNs as one of the most accurate state of the art monotonic networks \cite{kimScalableMonotonicNeural2023}, reportedly showing better results than regular DLNs in several regression tasks, making the current comparison very relevant.

We also wanted to compare performance to non-SQR settings. For this purpose, we use two more LSTM-Linear models, one point predictor trained with mean absolute loss and a QR with a predefined set of 11 quantiles. These are intended to show potential trade-offs made by using a probabilistic model with modeling capacity reserved to represent the implicit CDF instead of a set of discrete quantiles.

Lastly, in solar irradiance forecasting, smart persistence is seen as the most consistent baseline forecaster, and is used as a performance validation for more sophisticated models \cite{vandermeerReviewProbabilisticForecasting2018}. Smart persistence assumes that over a short time, irradiance is not changing significantly, and combines it with future information about irradiance under clear sky conditions. If a model cannot at least match the accuracy of smart persistence, it is considered ill-adjusted.
\begin{equation}\label{spersistance}
    f(t) = \frac{x(t)}{o(t)} * o(t+h) 
\end{equation}
With $x(t)$ being observed irradiance, and $o(t)$ being clear sky irradiance, we estimate future irradiance (in $h$ steps) based on the difference in observed and clear irradiance at $t$. In our condition of forecasting 36 hours into the future, we use the last day (i.e., 24 hours) as our source of information for the next one-and-a-half days. First, the previous 24 hours predict the next 24 hours.
Then, the remaining 12 hours of the forecast are predicted by the first half (12 hours) of the 24 hours of the previous day. This way, each timestep is predicted by a previous day's timestep at the same time of day.

\subsubsection{Metrics}
All models are evaluated following many of the recommendations for probabilistic solar irradiance forecasting evaluations \cite{lauretVerificationSolarIrradiance2019}. 
Further, it should be noted that all metrics are calculated as the average across the length of the complete forecast horizon. This averaging is omitted from the equations for brevity.

First, we evaluate the point prediction metrics mean absolute error (MAE), root mean squared error (RMSE), and skill score (SS), which are evaluated on the 0.5 quantile values. SS is the fraction between $MSE_f$ of the LSTM-DLN and $MSE_p$ from a smart persistence model, such that:
\begin{equation}\label{eq:skillscore}
    SS = 1 - \frac{MSE_f}{MSE_p}
\end{equation}

Further, we examine continuous rank probability scores (CRPS) as a probabilistic representation of distributional accuracy. CRPS is usually computed in continuous form as follows,
\begin{equation}\label{eq:crps}
    CRPS = \frac{1}{N}\sum^{N}_{i=1}{\int_{-\infty}^{+\infty}(\hat{F}_i(x) - 1_{x \gamma_i})^2 dx}
    \end{equation}
In our case, we employ an approximated version of CRPS computed by the sum of pinball loss over all evaluated quantiles \cite{narayanExpectedPinballLoss2023}:
    \begin{equation}\label{eq:approx_crps}
 CRPS =\frac{1}{N} \sum^{N}_{i=0} l_\tau(\gamma,\gamma')
\end{equation}
Examining the coverage properties of the CDF, we plot prediction interval coverage percentage (PICP) and reliability diagrams in addition to computing the average coverage error (ACE) score. 
PICP is a percentage of interval points correctly captured in each predication interval $c$ between the points $[\hat{I}_{0.5-\frac{c}{2}}, \hat{I} _{0.5+\frac{c}{2}}]$, which are the pair of opposing quantiles. PICP is better the closer it is to the target percentage coverage of its respective interval.

\begin{equation}\label{eq:picp}
    PICP_c = \frac{1}{N} \sum^N_{i=1}{s_i}, s_i =
\begin{cases}
1, &\text{if}\; \gamma_i  \in [\hat{I}_{0.5-\frac{c}{2}}, \hat{I} _{0.5+\frac{c}{2}}] \\
0, &otherwise
\end{cases}
\end{equation}
ACE is the average difference of all PICP scores (and thus all intervals $C$) and the target percentage coverage of their respective interval.
\begin{equation}\label{eq:ace}
    ACE = \frac{1}{|C|} \sum_{c \in C}{|c - PICP_c|}
\end{equation}
Lastly, reliability diagrams are calculated just as PICP, but instead of counting values between intervals, we count values correctly captured within a quantile level. 

\subsection{Implementation}

Model training, optimization, and testing are done on a system with an AMD EPYC 7702 64-Core CPU and an Nvidia RTX3090 GPU. 
All code is written in Python with PyTorch \cite{pytorch} and the package pytorch-lattice \cite{Williampytorchlattice2024}.

All models were optimized for hyperparameters in a sequential (coarse-to-finer) grid search with subsequent manual adjustment and visual inspection of probability distribution fit. General performance was assessed with CRPS and ACE, but additional metrics were also taken into account. A detailed description of hyperparameter optimization is given in Appendix A, while optimized model parameters are detailed in Appendix B.
Some hyperparameters were kept constant for brevity and based on experience in performance development.
 Constant hyperparameters are a batch size of 64 and the Adam optimizer. All probabilistic models were trained using pinball loss from equation \ref{eq:pinball_loss}.


\section{Results}

\begin{table*}[ht]
\centering
\begin{tabular}{l|c|c|c|c|c}
\hline
Models & CRPS & MAE & RMSE & ACE & SS \\
\hline
SP & - & 52.903 &96.472  & - & - \\
LSTM-PP & - & 43.152(0.336) & 74.933(0.602) & - &0.397(0.019)  \\
LSTM-QR & 30.506(0.083) &43.875(0.226)  &74.949(0.343)  & 0.052(0.002) & 0.396(0.011) \\
SQR: LSTM-Lin & 24.427(0.247) &45.556(0.360)  & 77.165(0.414) &0.173(0.003)  & 0.360(0.014) \\
SQR: LSTM-NN &27.690(0.099)  & 42.612(0.525) & 72.930(1.009) &0.075(0.003)  & 0.429(0.032)  \\
SQR: LSTM-CLin &25.920(0.184)  & 44.619(0.294) & 75.472(0.397) & 0.190(0.003) & 0.388(0.013) \\
SQR: LSTM-SMNN &29.333(0.657)  & 47.548(0.974) & 81.731(1.691) &0.079(0.657)  & 0.282(0.059) \\
SQR: LSTM-DLN & 29.175(0.216) & 43.266(0.534) & 74.520(0.850) & 0.061(0.003) &0.403(0.027) \\ 
\hline
\end{tabular}
\caption{Mean model performance comparison over 5 training runs with a different random seed. Standard deviation is displayed in brackets. SP is smart persistence, PP is the point predictor, and QR is regular quantile regression. SQR models use simultaneous quantile regression with an LSTM as $f_1$ and different $f_2$. L is a linear layer, NN is a non-linear neural network, CL is a constrained linear layer, SMNN is a scalable constrained neural network, and DLN is a deep lattice network. }\label{tab:performance}
\end{table*}

\begin{figure}[ht]
    \centering
\includegraphics[width=1.0\columnwidth]{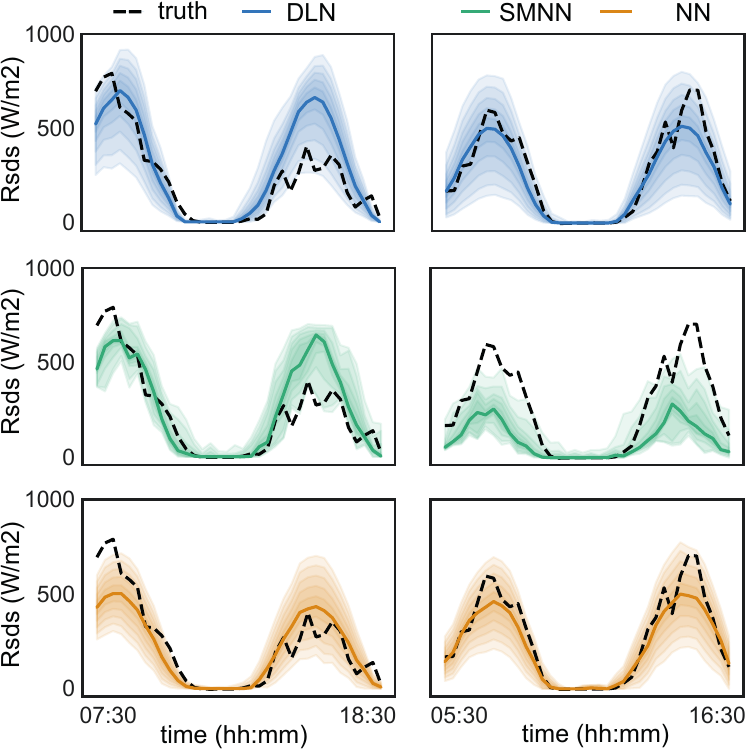}
\caption{Two 36h forecasts of the test data from LSTM-DLN (row 1), LSTM-SMNN (row 2), and LSTM-NN (row 3). Shown are 11 quantiles between 0.025 and 0.975.  Axis labels are consistent between subplots.}
\label{fig:comparison}
\end{figure}

Metric performance observations based on the test dataset are shown in table \ref{tab:performance}. Scores of the models are based on the average metric value over a horizon of 36 hours ahead.
All metrics displayed are better when smaller, except SS, which is better when larger. Results are the average performance over five optimization runs with different random seeds. In parentheses, the standard deviation of the metric is shown.
Generally, all models display positive SS and thus are better forecasters than the standard baseline.
In MAE and RMSE, the unconstrained LSTM-NN and our LSTM-DLN are not significantly different. The point predictor (LSTM-PP) is also not significantly different from LSTM-NN and LSTM-DLN in MAE, but shows a significant performance decline in RMSE compared to LSTM-NN. SQR approaches showing better performance may be due to regularization caused by representing all quantiles. 
In the ACE score, LSTM-QR performs best, with the second best being the LSTM-DLN. CRPS results are more complex. Linear and constrained linear appear to be the best performers. However, this is due to the way CRPS is approximated here (equation \ref{eq:approx_crps}). In pinball loss (equation \ref{eq:pinball_loss}), any value of $\gamma'$ close to $\gamma$ will result in lower loss, no matter the quantile value. As such, this loss can favour a distribution in which all quantiles have the same value close to the median of the target values. In both linear conditions, the CDF is observed to be small in width. This is why their respective ACEs are much worse than all others and appear to be ill-calibrated. Their performance in MAE and RMSE is similar to the non-linear competitors (LSTM-NN, SMNN, DLN), likely due to the networks being trained to focus on the median quantile. Due to this ill-calibration, they are generally performing worse. 

LSTM-SMNN displays slightly worse performance than LSTM-DLN in MAE and RMSE, while not being significantly different in CRPS and ACE. In ACE, this is the case, as its ability to form a wide and well-calibrated CDF appears to be inconsistent across random seeds. Even for well-adjusted examples, LSTM-DLN displays better results.

Figure \ref{fig:comparison} shows two different full 36h forecasts from NN, SMNN, and DLN. All plots in the same column are from the same timestamp. We can observe that the fit is not perfect for either of them, as the forecasting of changes in irradiance from one day to the next is not trivial. DLN shows the largest CDF width, which seems to help capture observations within its CDF. DLNs also seem to display a more skewed and non-Gaussian distribution compared to NN. SMNN shows more erratic behavior and poorer fit, reflected by its worse performance metrics.  

\begin{figure}[ht]
    \centering
\includegraphics[width=1.0\columnwidth]{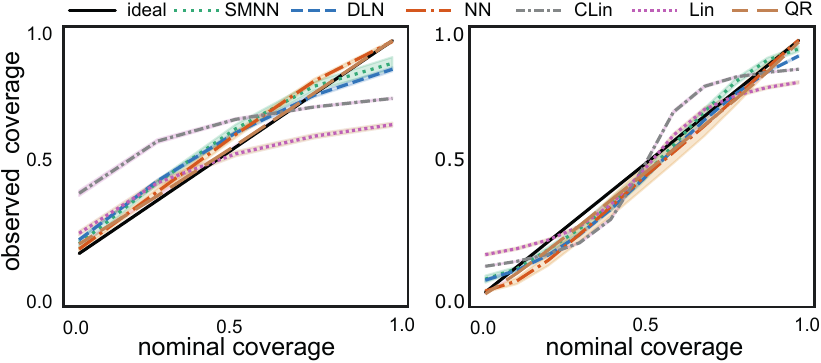}
\caption{PICP (left) and reliability diagram (right). PICP starts at approx. 0.2 as that is the smallest interval around the median. The values are means across all 5 models, with standard deviations displayed as colored areas around the curve.}
\label{fig:picps}
\end{figure}

In figure \ref{fig:picps}, we show CDF coverage performance for all models across intervals and quantiles. Ideal performance is marked by the black bar. Apart from constrained linear (CLin) and linear (Lin), all models seem to be close to the ideal. This echoes the previous observation that both linear approaches are ill-calibrated even though their median fit is good. In terms of interval coverage, the smaller intervals closer to the median show slight overestimation, while the lateral intervals do not always include all extreme values. 
In addition to mistakes in the extreme values, the reliability diagram shows quantiles between 0.2-0.5 to be more underestimated. This likely means that even the better-performing models make mistakes in the median of the CDF, leading to less correct values within the lower and higher quantiles around the median, which is not captured by the interval.

LSTM-NN, as the most well-performing competitor to our LSTM-DLN, is unconstrained with respect to quantiles. The overall average percentage of quantile crossovers displayed by the LSTM-NN over the whole test set and all quantiles is 3.075 \% with a standard deviation of 0.0744. On the other hand, DLNs are still much more complex in terms of parameter numbers than other networks. The DLN implementation has more than twice the number of parameters as SMNN  and takes more than 2s for every forward pass when run on a local laptop. Exact times and trainable parameter numbers for all models can be seen in Appendix C.

\section{Discussion}

This work aimed to show the effectiveness of using a DLN as a CDF multi-horizon forecaster. We do this by comparing against unconstrained approaches, as they are more flexible, thus should perform better, and a state of the art monotonic network approach, which showed better results than DLNs in a different application context.

The results show that our adaptation has similar performance to the unconstrained alternative, but shows better coverage in its quantiles. In conclusion, the expected limited expressiveness in the DLN induced by monotonic constraints is successfully mitigated by our adaptations. Our DLN adaptation also shows higher performance than an SMNN-based implementation, highlighting the competitiveness of this approach for time series forecasting within the state of the art of monotonic networks. Further, we show the necessity of a non-linear forecaster by comparing it against a linear representation of the implicit CDF, as LSTMs with both types of linear output layers display notably worse calibration compared to all other models. Well-performing SQR approaches do not show many tradeoffs to point predictor or predetermined QR. QR shows an edge in coverage performance; however, the other top-performing models also have very good coverage. 

Overall, our approach represents a successful way to apply DLNs on multi-horizon time series forecasting of CDFs. It should be highlighted that this implementation is highly modular, and we would encourage further investigation with other embedding models, more comparisons across quantile modelling techniques, or the usage of the DLN output layer in different contexts. In its current implementation, the SQR-based DLN can be considered useful for highly detailed modeling of a variable-sized (partial) inverse CDF. Contrary to approaches outside of monotonic networks, it can also be used to constrain more features (than just quantiles) monotonically. Lastly, an implementation that does not need further embedding before applying a Lattice may benefit from its increased explainability. In contrast to neural networks, lattices are interpolated look-up tables that model the whole range of possible outputs. As such, relations between input and output features are not hidden in a black-box and may even be plotted.

Nevertheless, DLNs still suffer from a very high complexity in the number of trainable parameters and execution time, the decrease of which remains an ongoing research question. 
\section{Conclusion}
In this work, we show that monotonous network research can be successfully connected with probabilistic forecasting applications. DLNs can be applied to time series forecasting, yielding a complete, implicit CDF with performance comparable to that of the unconstrained alternative and surpassing the state of the art monotonic network competitor.

\section{Acknowledgements}
We would like to thank our friends and colleagues at the Department of Technology Systems and the Institute for Energy Technology for the (ongoing) helpful discussions and support. We would also like to thank the anonymous reviewers for their insightful feedback and enriching comments on the draft of this work. 

\appendix

\makeatletter
\@ifundefined{isChecklistMainFile}{
  \newif\ifreproStandalone
  \reproStandalonetrue
}{
  \newif\ifreproStandalone
  \reproStandalonefalse
}
\makeatother

\ifreproStandalone
\documentclass[letterpaper]{article}
\usepackage[submission]{aaai2026}
\setlength{\pdfpagewidth}{8.5in}
\setlength{\pdfpageheight}{11in}
\usepackage{times}
\usepackage{helvet}
\usepackage{courier}
\usepackage{xcolor}
\usepackage{natbib}
\frenchspacing

\newcommand{\citeauthorandyear}[1]{\citeauthor{#1} (\citeyear{#1})}

\begin{document}

\fi
\setlength{\leftmargini}{20pt}
\makeatletter\def\@listi{\leftmargin\leftmargini \topsep .5em \parsep .5em \itemsep .5em}
\def\@listii{\leftmargin\leftmarginii \labelwidth\leftmarginii \advance\labelwidth-\labelsep \topsep .4em \parsep .4em \itemsep .4em}
\def\@listiii{\leftmargin\leftmarginiii \labelwidth\leftmarginiii \advance\labelwidth-\labelsep \topsep .4em \parsep .4em \itemsep .4em}\makeatother

\setcounter{secnumdepth}{1}
\renewcommand\thesubsection{\arabic{subsection}}
\renewcommand\labelenumi{\thesubsection.\arabic{enumi}}

\newcounter{checksubsection}
\newcounter{checkitem}[checksubsection]

\newcommand{\checksubsection}[1]{%
  \refstepcounter{checksubsection}%
  \paragraph{\arabic{checksubsection}. #1}%
  \setcounter{checkitem}{0}%
}

\newcommand{\checkitem}{%
  \refstepcounter{checkitem}%
  \item[\arabic{checksubsection}.\arabic{checkitem}.]%
}
\newcommand{\question}[2]{\normalcolor\checkitem #1 #2 \color{blue}}
\newcommand{\ifyespoints}[1]{\makebox[0pt][l]{\hspace{-15pt}\normalcolor #1}}
\appendix
\ifdefined\HIDEDOCB

\label{appendix a}

\label{appendix b}

\label{appendix c}

\else

\section{ Hyperparameter Optimization}\label{appendix a} 
\begin{table}[ht]
    \centering
    \begin{tabular}{c|c}
    \hline
    Hyperparameter &  Range\\
    \hline
        learning rate &  0.1 - 1e-7 in 0.1 increments\\
        layers & 1 - 4  \\
        units  & 32 - 512 in base 2 increments 
    \end{tabular}
    \caption{LSTM - hyperparameter optimization ranges}
    \label{tab:LSTM-HO}
\end{table}

The large number of hyperparameters resulted in a potentially large search space. Therefore, an iterative grid search, with selectively decreasing or increasing the grid space, was used to find the optimal hyperparameters. A final set was then examined through separate validation runs, whereby all evaluation metrics and visual inspection of results were used to determine optimal settings. 

As the LSTM-DLN approach is divided into 2 models, $f_1$ and $f_2$, we optimized them separately. 
We started the optimization by optimizing the LSTM with the unconstrained NN output layer (so LSTM-NN). We set the NNs architecture to use the same number of units as the LSTM, with 2 layers and one ReLU activation function in between. Any such network is a universal function approximator and should be able to model quantile distributions. Optimized LSTM hyperparameters based on the LSTM-NN performance can be examined in Table \ref{tab:LSTM-HO}.

Linear, constrained Linear, the LSTM point predictor, and QR did not need further hyperparameter optimization as they all had a linear layer as the output layer $f_2$.

SMNN needs optimization of its three distinct parallel layer architecture. Please refer to \citeauthorandyear{kimScalableMonotonicNeural2023} for formal definitions of the SMNN.
In summary, these layers are named the ReLu layer, the Confluence unit, and the Exponentiated unit. The ReLU layer takes the unconstrained values and propagates them to the next ReLU layer and the next Confluence unit. The Confluence unit takes the unconstrained values, or input from the last ReLU layer, and propagates them to the next Exponentiated unit. In turn, the exponentiated unit takes constrained values when in the first layer. Input of the Exponentiated unit in the second layer is given by the output of the last Confluence unit and the previous Exponentiated unit. All parallel layers can be cascaded, so the number of layers can be varied freely. In the original paper, two layers are used across all experiments, so we do the same here. The activation of all three parallel layers is finally concatenated and fed into a fully connected layer for a final output.
Hyperparameter-wise, we validate for both layers in all 3 parallel layer structures separately over the range 31 - 1024 parameters in increments of exponents of 2. 

For the adapted DLN, we validated the keypoints of the higher-dimensional lattice and calibration layers, and the lattice input size. Similarly to \citeauthorandyear{narayanExpectedPinballLoss2023}, we adjust calibration keypoints related to quantile values separately to ensure more expressiveness. The validated hyperparameter ranges can be examined in Table \ref{tab:DLN-HO}.
 Calibration keypoints will influence the expressiveness of their input feature on the result, while the lattice influences the expressiveness of the interaction between the different features that are given as its input. 
 
Lattice keypoints $k$ and input sizes $D$ are limited by the $k^D$ complexity. While needing space for the large output of the LSTM (which was optimized to 128 units), results can quickly result in too large weight matrices for the GPUs we had access to. The worst case here, $21^4 * (128/4)$, results in over 6 million parameters just for the lattice ensemble.

\begin{table}[t]
    \centering
    \begin{tabular}{c|c}
    \hline
    Hyperparameter &  Range\\
    \hline
        Calibration keypoints &  2-100 in increments of 10\\
        Calib. quantile keypoints & 2-100 in increments of 10 \\
        Lattice keypoints  & 2,5,11,21 \\
        Output Calib. keypoints & 2-100 in increments of 10 \\
        Lattice input size & 2,3,4
    \end{tabular}
    \caption{DLN - hyperparameter optimization ranges}
    \label{tab:DLN-HO}
\end{table}

It is noteworthy that model flexibility/complexity and number of training epochs facilitate a trade-off between probabilistic coverage and median fit of the prediction. Quantile loss averaged across quantiles can minimize itself below a minimized mean absolute error. This happens when all quantiles are closer to the median than their target quantiles. In these cases, the network likely has a too narrow CDF. In other words, the network is ill-calibrated. As such, the network needs to be optimized for both quantile loss and calibration metrics.
Thus, during the hyperparameter optimization, results were examined not only for CRPS but also for ACE. 

Across all SQR networks, except the DLN, we examined the tendency to fail at widening their respective probability distribution, resulting in a worse ACE. To combat this, we initially set higher learning rates and lowered them with a scheduler after a couple of steps. We tested for an epoch-based scheduler and a scheduler depending on validation loss changes. The exact rate of the scheduled learning rate decrease is examined in Appendix \ref{appendix b}. As a side effect of using the scheduler, epoch numbers needed to reach optimal values in training decreased. This was overall beneficial, especially in the case of DLNs, which have been observed to take 10 times as many training steps as NNs otherwise \cite{narayanExpectedPinballLoss2023}.

\section{Optimized models}\label{appendix b} 
Optimal hyperparameters for the LSTM embedding model were as shown in Table \ref{tab:lstm-OHP}, with a constant batchsize of 64, Hiddensize of 128, and Layernumber of 2.
\begin{table}[ht]
    \begin{tabular}{lccp{2.6cm}}
        \hline
        Models & Epochs& Learning Rate  & Scheduler \\
        \hline
        NN &30& 0.001 & 0.1 step if CRPS increases.\\
        PP &10& 0.001 & 0.1 step at epoch 1,2,3 and if CRPS increases for 1 epoch. \\
        Lin &20& 0.001 & 0.1 step if CRPS increases for 2 epochs\\
        CLin  &300& 0.1 & 0.01 step at epoch 1,2 and 0.1 step if CRPS increases.\\
        QR   &250& 0.001 &0.1 step at epoch 1,2,3 and if CRPS increases for 1 epoch.  \\
        DLN    &10& 0.001 & 0.5 step at epoch 1,2,3,4 \\
        SMNN     &20& 0.001 & 0.1 step if CRPS increases for 2 epochs \\
        \hline
        
    \end{tabular}
    \caption{Varying hyperparameters unrelated to different model architectures. Scheduler steps are multiplication factors on the initial learning rate. CRPS is calculated on the validation data.}\label{tab:lstm-OHP}
\end{table}

Epochs slightly varied as early stopping was used to prevent overfitting.

Optimal hyperparameters for the DLN model were as follows:
\begin{itemize}
    \item Calibration keypoints: 61
    \item Calib. quantile keypoints: 11
    \item Lattice keypoints: 21
    \item Output Calib. keypoints: 61
    \item Lattice input size: 2
\end{itemize}

Optimal hyperparameters for 3 parallel layers of the SMNN model were as follows:
\begin{itemize}
    \item exponentiated unit: (128, 128) 
    \item confluence unit: (128, 64)
    \item ReLU unit: (128, 256)
\end{itemize}

\section{Extended Results}\label{appendix c}

\begin{table}[ht]
\centering

\begin{tabular}{l|c|c}

\hline
Models & Time(s) & Parameters\\
\hline
SP & - & - \\
LSTM-PP & 0.002(0.000) &329,252 \\
LSTM-QR & 0.011(0.002) & 375,692 \\
SQR: LSTM-Lin & 0.006(0.001) & 329,288 \\
SQR: LSTM-NN &0.010(0.001) & 346,058 \\
SQR: LSTM-CLin &0.006(0.001) &329,288 \\
SQR: LSTM-SMNN &0.059(0.002) &448,228  \\
SQR: LSTM-DLN & 2.604(0.092)&931,375\\
\hline
\end{tabular}
\caption{Additional Measures: Average time (Variance in brackets) in seconds of one forward pass of algorithm 2 with an AMD RYZEN 7 7735HS, an RTX 4070 laptop GPU, and 16GB of RAM. Parameters are the total number of parameters in the models.}\label{tab:mess}
\end{table}

We show parameters needed and time spent for one forward pass for 11 quantiles in table \ref{tab:mess}. Here we can see that DLN is much more complex and time-intensive compared to regular neural networks. SMNN also displays this disparity, but to a much lesser degree. 

\ifreproStandalone
\end{document}
\fi
\fi

\bibliography{Paper2}

\end{document}